\documentclass{article}

\usepackage{arxiv}

\usepackage[utf8]{inputenc} % allow utf-8 input
\usepackage[T1]{fontenc}    % use 8-bit T1 fonts
\usepackage{hyperref}       % hyperlinks
\usepackage{url}            % simple URL typesetting
\usepackage{booktabs}       % professional-quality tables
\usepackage{amsfonts}       % blackboard math symbols
\usepackage{nicefrac}       % compact symbols for 1/2, etc.
\usepackage{microtype}      % microtypography
\usepackage{graphicx}
\usepackage[numbers]{natbib}
\usepackage{doi}

\usepackage{float}
\usepackage{placeins}
\usepackage{caption}

\usepackage{array}

\usepackage{xurl}

\newcolumntype{L}[1]{>{\raggedright\arraybackslash}p{#1}}

\providecommand{\undertitle}{}

\title{Who Became Financially Vulnerable After COVID-19? A Population-Level Machine Learning Analysis Using MEPS Data}

\author{
\textbf{Alexey Kresin}$^{1,5,*}$,
\textbf{Zien Cheng}$^{2,5}$,
\textbf{Ammar Ahad}$^{5}$,
\textbf{Ebiyomare Kelvin}$^{3,5}$,\\
\textbf{Manish Sivaratri}$^{4,5}$,
\textbf{Prabhjeet Singh}$^{5,6}$,
\textbf{Omar Aljawfi}$^{5,6}$,\\
\textbf{Olabisi Ojo}$^{7,5}$,
\textbf{Nawar Shara}$^{5,6}$\\[0.8em]
\texttt{forkresin@gmail.com}\\
\textit{* Corresponding Author}\\[0.8em]
\small
$^{1}$ Hood College, Frederick, MD, USA\\
$^{2}$ Georgetown University, Washington, DC, USA\\
$^{3}$ University of Benin, Benin City, Nigeria\\
$^{4}$ Department of Computer Science and Engineering, NIST University, Odisha, India\\
$^{5}$ AI CoLab: MedStar--Georgetown Collaborative Center for Artificial Intelligence\\
in Healthcare Research and Education, Washington, DC, USA\\
$^{6}$ MedStar Health Research Institute, Washington, DC, USA\\
$^{7}$ Department of Natural Sciences, Albany State University, Albany, Georgia, USA
}

\hypersetup{
pdftitle={Who Became Financially Vulnerable After COVID-19? A Population-Level Machine Learning Analysis Using MEPS Data},
pdfauthor={Alexey Kresin, Zien Cheng, Ammar Ahad, Ebiyomare Kelvin, Manish Sivaratri, Prabhjeet Singh, Omar Aljawfi, Olabisi Ojo, Nawar Shara},
pdfkeywords={healthcare financial burden, MEPS, COVID-19, machine learning, temporal generalization, health disparities},
}

\begin{document}
\maketitle

\begin{abstract}
Financial strain associated with health care continues to be a persistent concern in the United States and was likely affected by COVID-19 pandemic-related disruptions. This study analyzed patterns of financial vulnerability before and after the pandemic based on Medical Expenditure Panel Survey (MEPS) data for 2019 and 2021. We defined high financial burden as out-of-pocket healthcare spending greater than 10\% of family income and compared differences by demographic and socioeconomic groups using survey-weighted estimates to allow for nationally representative interpretation.

Descriptive subgroup analysis was used alongside predictive modeling to provide further insight into factors associated with financial vulnerability. Interpretable logistic regression models were applied to estimate odds ratios. Machine learning methods such as random forests and gradient boosting were used to assess predictive performance. We assessed the temporal stability of the learned relationships by performing temporal generalization experiments, training the models on pre-pandemic data and testing them on post-pandemic observations.

Our results indicate that financial vulnerability is strongly related to income level, insurance status, and prescription drug spending. The subgroup analyses show that differences across population groups persist, with some evidence of increased burden in previously vulnerable populations during the post-pandemic period. However, despite these differences, models trained on pre-pandemic data exhibit only modest differences in performance when applied to post-pandemic data, suggesting that fundamental relationships are relatively stable.

In conclusion, our study provides a population-level picture of healthcare financial vulnerability during the COVID-19 period, underscoring the persistence of inequities and the relative stability of predictive relationships. These results may serve to inform future risk identification and decision support work, while also emphasizing the ongoing salience of vulnerable populations.
\end{abstract}

% keywords can be removed

\section{Introduction}

Health care costs are a major public health and policy concern in the United States. Even for individuals with insurance, high out-of-pocket medical costs can impose a large financial burden, as cost-sharing requirements such as co-payments, deductibles, and coinsurance have increased significantly over the past two decades, shifting healthcare costs directly onto patients \cite{claxton2024costsharing}. This is especially burdensome for low-income individuals, those with chronic conditions, and those with elevated healthcare needs, including older adults, disabled individuals, and patients requiring specialist care or long-term medication use. Prior research has found that health-care-related financial strain can lead to delayed care, lower medication adherence, medical debt, and greater financial instability \cite{baird2016financial,sheng2018medical}. These issues became urgent during and after the COVID-19 pandemic. It disrupted access to medical services, jobs, insurance, as well as changed healthcare usage across the country.

The COVID-19 pandemic disrupted the healthcare system and broader economy, potentially changing patterns of financial vulnerability \cite{nam2023covid,hill2023healthcare,zuvekas2021impacts}. Employment instability and reduced healthcare use in the early pandemic period, and existing socioeconomic and demographic disparities, may have widened in the post-pandemic period. Understanding which populations remained financially vulnerable, and whether predictive relationships changed over time, is important for public health monitoring and future decision support efforts.

In this study, we examine healthcare-related financial vulnerability using the Medical Expenditure Panel Survey (MEPS) Full-Year Consolidated files from 2019 and 2021, representing pre- and post-pandemic periods. We defined high financial burden as out-of-pocket healthcare spending exceeding 10\% of family income, a commonly used threshold in studies of catastrophic health expenditure and healthcare affordability \cite{baird2016financial,getachew2023threshold}. The analysis followed a repeated cross-sectional design and combined descriptive subgroup analysis with predictive modeling approaches. Survey-weighted estimates were used for subgroup analyses to support nationally representative interpretation.

Our modeling framework included interpretable logistic regression models to estimate associations between predictors and financial burden, alongside machine learning methods such as random forests and gradient boosting to evaluate predictive performance. To reduce the risk of target leakage, variables directly used in the construction of the outcome, including total out-of-pocket spending, family income, and burden-related measures, were excluded from model features. We also conducted temporal generalization experiments to evaluate whether models trained on pre-pandemic data maintained predictive performance when applied to post-pandemic observations.

This study makes several contributions. First, it examines healthcare financial vulnerability pre- and post-COVID-19 pandemic using nationally representative MEPS data. Second, it compares differences across demographic, income, and insurance subgroups using survey-weighted estimates. Third, it evaluates temporal stability of predictive relationships with cross-period generalization experiments, to assess whether pre-pandemic patterns remain informative in the post-pandemic setting.

The study does not attempt to identify causal effects of the pandemic. Instead, it emphasizes the identification of patterns, associations, and the constancy of predictive relationships over time. The findings contribute to current discussions around healthcare affordability and financial vulnerability and point to the potential for predictive analytics to help identify populations at increased financial risk.

\section{Related Work}

Researchers have widely studied healthcare costs and healthcare-related financial strain in the United States, particularly in low-income populations, the uninsured, and patients with chronic conditions. High out-of-pocket healthcare costs can lead to several adverse effects, including delayed care, lower medication adherence, medical debt, and greater financial instability \cite{baird2016financial,sheng2018medical}. Researchers often assess healthcare financial burden by comparing healthcare spending to household income, which helps identify groups facing disproportionate financial strain \cite{baird2016financial,underinsurance2024}.

The COVID-19 pandemic disrupted the healthcare system and broader economy, leading to increased attention toward healthcare-related financial vulnerability. Previous research has investigated changes in healthcare utilization, health insurance coverage, employment instability, and healthcare spending during the pandemic period \cite{zuvekas2021impacts,jacobs2023changes}. Early pandemic conditions were associated with declines in routine healthcare utilization, delays in care, and greater economic uncertainty \cite{thomas2021forgone,korenman2023health}. Although numerous studies have examined the short-term healthcare and economic disruptions associated with COVID-19, whether patterns of healthcare financial vulnerability persist into the post-pandemic period remains an active area of research.

Nationally representative survey datasets, such as the Medical Expenditure Panel Survey (MEPS) \cite{meps2024home,meps_public_use_files}, are commonly used to study healthcare expenditures, insurance coverage, healthcare utilization, and affordability in the United States. MEPS is particularly useful for population-based analyses of healthcare financial burden because it contains detailed information on healthcare expenditures, demographic characteristics, insurance coverage, employment, and socioeconomic factors. Previous studies using MEPS data have examined differences in healthcare burden across income groups, insurance categories, and populations with chronic conditions \cite{baird2016financial,dong2021drug}.

At the same time, predictive modeling methods have become increasingly common in public health analytics and health services research. Logistic regression remains widely used because of its interpretability and ability to estimate associations between predictors and binary outcomes through odds ratios. More recently, machine learning methods such as random forests and gradient boosting have been applied to healthcare prediction tasks due to their flexibility and ability to capture non-linear relationships and interactions among variables.

Despite growing interest in predictive healthcare analytics, relatively few studies have examined the temporal stability of predictive relationships across major societal disruptions such as the COVID-19 pandemic. Assessing the predictive performance of models trained on pre-pandemic data in post-pandemic settings may provide insight into the stability of healthcare financial vulnerability patterns over time.

In this study, we extend previous work by combining nationally representative subgroup analysis with interpretable statistical modeling and machine learning methods. In addition, our study explicitly evaluates temporal generalization between pre- and post-pandemic periods and removes variables directly involved in the construction of the outcome definition to reduce the risk of target leakage.

\section{Data}

\subsection{Data Source}

Our analysis used data from the Medical Expenditure Panel Survey (MEPS) Full-Year Consolidated files for 2019 and 2021 \cite{meps2019hc216,meps2021hc233}. MEPS is a nationally representative survey of the U.S. civilian non-institutionalized population that provides detailed data on health care utilization, expenditures, insurance coverage, demographic characteristics, and socioeconomic status. The 2019 file was used to represent the pre-pandemic period while the 2021 file was used to represent the post-pandemic period. The final long-format analytical dataset included 54,990 observations across the two study years, with 27,682 observations from 2019 and 27,308 observations from 2021.
In this paper, 2021 is used as the post-2020 comparison period relative to the initial pandemic shock, while recognizing that pandemic-related disruptions continued into 2021.

\subsection{Study Design}

The analysis used a repeated cross-sectional design. Individuals observed in 2019 and 2021 were treated as separate cross-sectional samples, not the same individuals followed over time. Thus, the study was a comparison of population-level patterns before and after the COVID-19 pandemic, not an estimation of individual-level longitudinal change. With this design, we could investigate whether the subgroup patterns and predictive relations looked similar or different in the two time periods.

\subsection{Outcome Definition}

The main outcome was high financial burden in healthcare. We defined high financial burden as out-of-pocket health care spending exceeding 10\% of family income, consistent with commonly used expenditure-to-income thresholds in healthcare affordability research. This threshold was meant to flag situations where the household’s health expenses are a significant burden on its resources. The outcome variable was coded as 1 if the individual exceeded this threshold and 0 otherwise.

\subsection{Feature Set}

The predictors included demographic, socioeconomic, insurance, and health variables available in the MEPS Full-Year Consolidated files. These variables were used to describe individuals and to estimate the association with high financial burden. Examples included age, sex, race and ethnicity, poverty status, insurance coverage and measures of healthcare need or utilization.

To minimize the risk of target leakage, model features excluded variables used directly in building the outcome. Such variables included total out-of-pocket health care spending, family income, the calculated burden ratio, and any other burden-related variables calculated from these quantities. This exclusion helped ensure that the predictive models did not have direct access to the information defining the target variable.

\subsection{Variable Selection and Feature Inclusion}

To improve transparency, we categorized the variables used in this study according to their role in the analytical pipeline. For transparency and reproducibility, all variables considered during the study were categorized into four groups according to their role in the analytical pipeline: (1) variables used to define the outcome, (2) candidate predictor variables available in the MEPS Full-Year Consolidated files, (3) variables included in the final predictive models, and (4) survey design variables used only for weighted descriptive analyses.

Candidate predictors were selected a priori based on domain knowledge, variable availability, and their potential applicability for estimating healthcare financial vulnerability. First, the variable had to be available in both the 2019 and 2021 MEPS Full-Year Consolidated files. Second, it needed to represent demographic, socioeconomic, insurance-related, employment, or healthcare characteristics that could plausibly be available when estimating financial vulnerability. Third, variables directly involved in constructing the outcome were excluded to prevent target leakage.

Table~\ref{tab:variable_groups} summarizes the role of each variable category in the analytical framework.

\begin{table}[H]
\centering
\caption{Variable groups used in the analytical framework.}
\label{tab:variable_groups}
\begin{tabular}{p{4cm}p{6.5cm}p{4cm}}
\toprule
\textbf{Variable Group} & \textbf{Examples} & \textbf{Role} \\
\midrule

Outcome-defining variables &
Family income, out-of-pocket healthcare expenditures, burden ratio &
Used to construct the target variable; excluded from predictive models \\

Candidate predictor variables &
Demographic, socioeconomic, insurance, employment, and healthcare variables &
Considered for descriptive and predictive analyses \\

Final modeling variables &
Variables retained after leakage removal and preprocessing &
Used as predictors in logistic regression and machine learning models \\

Survey design variables &
MEPS person-level survey weights (PERWT) &
Used only for nationally representative subgroup estimates \\

\bottomrule
\end{tabular}
\end{table}

\begin{table}[H]
\centering
\caption{Variables considered for analysis and their role in the analytical framework.}
\label{tab:variables}
\small
\begin{tabular}{L{2.8cm} L{4.5cm} L{2.2cm} L{5.3cm}}
\toprule
\textbf{Variable} &
\textbf{Description} &
\textbf{Role} &
\textbf{Reason} \\
\midrule

AGE &
Unified age variable created from AGE19X and AGE21X &
Predictor &
Demographic characteristic associated with healthcare utilization and financial vulnerability. \\

SEX &
Sex &
Predictor &
Basic demographic characteristic. \\

RACEV1X &
Race / Ethnicity &
Predictor &
Used to evaluate disparities across population groups. \\

POVCAT19 / POVCAT21 &
Poverty category &
Predictor &
Primary socioeconomic indicator of financial vulnerability. \\

INSURC19 / INSURC21 &
Insurance coverage &
Predictor &
Insurance status influences out-of-pocket healthcare costs. \\

RXEXP19 / RXEXP21 &
Prescription drug expenditures &
Predictor &
Proxy for healthcare need and medication burden. \\

EMPST\_FINAL &
Employment status &
Predictor &
Derived from multiple MEPS interview rounds to summarize employment status. \\

PROBPY\_final &
Problems paying medical bills &
Predictor &
Indicator of healthcare-related financial hardship. \\

CRFMPY\_final &
Could not afford medical care &
Predictor &
Captures barriers to healthcare affordability. \\

PYUNBL\_final &
Unable to pay medical bills &
Predictor &
Additional indicator of healthcare-related financial strain. \\

PERWT19F / PERWT21F &
Person-level survey weights &
Survey weighting &
Used only to obtain nationally representative subgroup estimates. \\

\midrule

FAMINC19 / FAMINC21 &
Family income &
Excluded &
Used directly to define the outcome variable. \\

TOTSLF19 / TOTSLF21 &
Out-of-pocket healthcare expenditures &
Excluded &
Used directly to define healthcare financial burden. \\

burden19 / burden21 &
Out-of-pocket spending divided by family income &
Derived outcome &
Continuous burden measure used to derive the binary outcome. \\

high\_burden19 / high\_burden21 &
Binary indicator of high financial burden &
Outcome &
Primary response variable used for statistical and machine learning analyses. \\

\bottomrule
\end{tabular}
\end{table}

\subsection{Data Preprocessing}

Before analysis, the combined 2019 and 2021 MEPS dataset underwent several preprocessing steps to improve data quality and ensure consistency across study years. Standard MEPS missing-value codes corresponding to nonresponse, inapplicable questions, and other missing data patterns were converted to missing values. Variables collected using different names across survey years were harmonized to create unified predictors. For example, age was represented by a single variable created from the year-specific age variables, while employment status and selected healthcare hardship indicators were derived by combining information collected across multiple MEPS interview rounds into unified variables.

To improve data quality, observations with non-positive family income were excluded because healthcare financial burden could not be meaningfully calculated for these records. In addition, households reporting family income greater than \$800,000 were excluded to reduce the influence of extreme outliers. Healthcare financial burden was calculated separately for 2019 and 2021 as the ratio of annual out-of-pocket healthcare expenditures to family income. Burden values were constrained to the interval $[0,1]$ to limit the influence of extreme ratios arising from very small denominators.

Finally, binary outcome variables were created by applying the predefined threshold for high healthcare financial burden. An individual was classified as having high financial burden if annual out-of-pocket healthcare expenditures exceeded 10\% of family income. The resulting processed dataset was then used for descriptive subgroup analyses, logistic regression, machine learning models, and temporal generalization experiments.

\begin{figure}[H]
    \centering
    \includegraphics[width=0.95\textwidth]{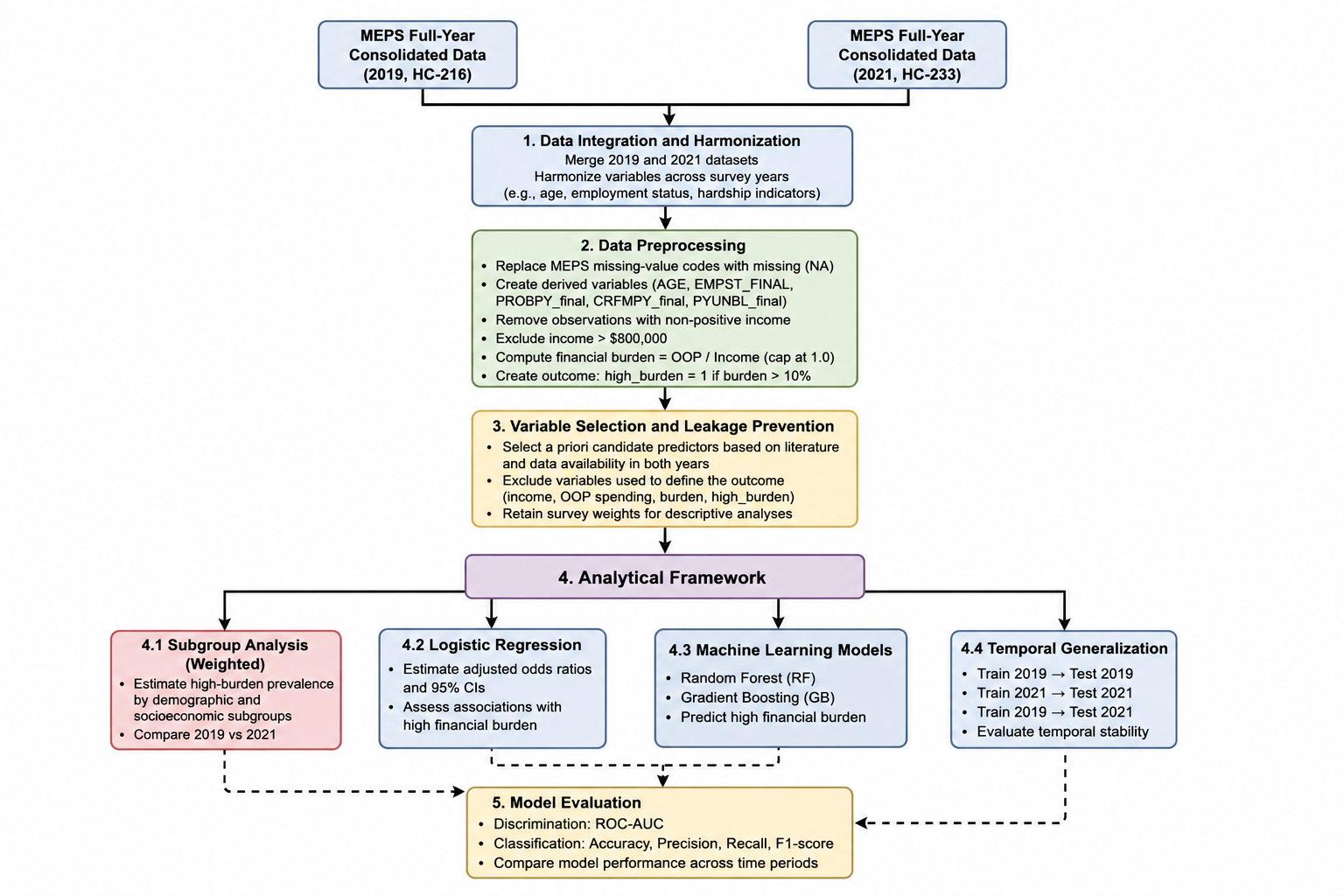}
    \caption{Overview of the study design and analytical workflow. MEPS data from 2019 and 2021 were harmonized, preprocessed, and used for subgroup analysis, statistical modeling, machine learning, and temporal generalization experiments.}
    \label{fig:workflow}
\end{figure}

\subsection{Survey Weights}

Survey weights were applied in the descriptive subgroup analysis to support nationally representative interpretation. Specifically, MEPS person-level weights were used to estimate the prevalence of high financial burden across demographic, socioeconomic, and insurance-related subgroups. These weighted estimates were used to account for the MEPS survey design and improve representation of the U.S. civilian non-institutionalized population.

Survey weights were not applied to the logistic regression or machine learning models. These analyses focused on adjusted associations, predictive performance, and temporal generalization within the analytical datasets rather than design-based population inference. Accordingly, nationally representative interpretation is limited to the survey-weighted descriptive subgroup estimates.

\section{Methods}

\subsection{Subgroup Vulnerability Analysis}

We first conducted a descriptive subgroup analysis to examine how healthcare financial burden differed across population groups in 2019 and 2021. Subgroups were defined using demographic, socioeconomic, insurance-related, and health-related variables available in the MEPS Full-Year Consolidated files. For each subgroup, we estimated the proportion of individuals classified as having high financial burden.

Survey weights were applied in this analysis to support nationally representative interpretation. Weighted estimates were calculated separately for 2019 and 2021, allowing comparison of subgroup burden rates before and after the COVID-19 pandemic. This analysis was intended to identify groups with higher observed financial vulnerability and to describe whether these patterns appeared to change across the two study periods.

\subsection{Logistic Regression}

Logistic regression was used as the primary interpretable statistical model. The outcome variable was the binary indicator for high financial burden. Predictor variables included demographic, socioeconomic, insurance-related, and health-related characteristics, while variables directly used to construct the outcome were excluded to reduce the risk of target leakage.

The logistic regression model estimated the adjusted association between each predictor and the odds of high financial burden, conditional on the other variables included in the model. Results were summarized using odds ratios and 95\% confidence intervals. Because the analysis was observational and cross-sectional, these associations should not be interpreted as causal effects.

The logistic regression model was fitted without survey weights because the objective was to examine adjusted associations within the analytical dataset rather than perform design-based population inference. After excluding observations with missing age or prescription expenditure values, the final analysis included 54,541 observations, of which 2,004 were classified as having high financial burden.

\subsection{Machine Learning Models}

In addition to logistic regression, we evaluated tree-based machine learning models to assess predictive performance. The models included random forest and gradient boosting classifiers. These methods were included because they could capture nonlinear relationships and interactions among predictors that may not be fully represented by a standard logistic regression model.

All models were trained using the same leakage-reduced feature set. The goal of the machine learning comparison was not only to identify the best-performing model, but also to examine whether more flexible models provided meaningful predictive improvement over logistic regression. This comparison helped balance interpretability and predictive performance in the context of healthcare financial vulnerability.

\subsection{Temporal Generalization Experiments}

We also ran temporal generalization experiments across the 2019 and 2021 MEPS datasets to assess the stability of predictive relationships over time. These experiments compared model performance across time periods and within the same time period.

The main temporal evaluation settings were:

\begin{enumerate}
    \item Training and testing within the 2019 dataset.
    \item Training and testing within the 2021 dataset.
    \item Training on the 2019 dataset and testing on the 2021 dataset.
\end{enumerate}

The first two settings measured the within-period predictive performance, and the third setting measured if the relationships learned from the pre-pandemic data remained informative to post-pandemic observations. A large decline in cross-period performance would suggest that predictive relationships changed substantially over time, while comparable performance would suggest relative temporal stability.

\subsection{Evaluation Metrics}

We assessed model performance with standard classification metrics. The primary metric was the area under the receiver operating characteristic curve (ROC-AUC), which summarized the model’s ability to discriminate between individuals with and without high financial burden across classification thresholds. ROC-AUC was useful in this setting because it is threshold-independent and allows for comparison across models and time periods.

Additional metrics like accuracy, precision, recall, and F1-score may provide a more complete view of classification performance. Threshold-dependent metrics must be interpreted with caution, as high financial burden may be relatively infrequent compared with the non-burden group. In particular, recall could be used to understand how well the model captured individuals with high financial burden, and precision could be used to understand the percentage of cases predicted to have high burden that are actually high burden.

\subsection{Leakage Prevention}

Target leakage was an important methodological concern in this study. Variables such as out-of-pocket health care expenditures and family income were not included in the predictive feature set because the outcome was defined using these variables and any directly derived measures of burden. Adding these variables would have allowed the model to indirectly reconstruct the outcome and would have resulted in overly optimistic estimates of performance.

This leakage prevention step was consistently applied for all modeling approaches including logistic regression, random forest, and gradient boosting. Thus, model performance reflected the predictive value of demographic, socioeconomic, insurance-related, and health-related characteristics and not direct access to the variables used to define the target.

\section{Results}

\subsection{Subgroup Differences in Financial Burden}

A subgroup analysis using survey weights showed that high healthcare financial burden was not evenly distributed among population groups. Higher burden rates were concentrated among economically vulnerable groups, those without insurance, and groups with indicators of greater need for health care in both study years.

\begin{figure}[H]
\centering
\includegraphics[width=0.75\linewidth]{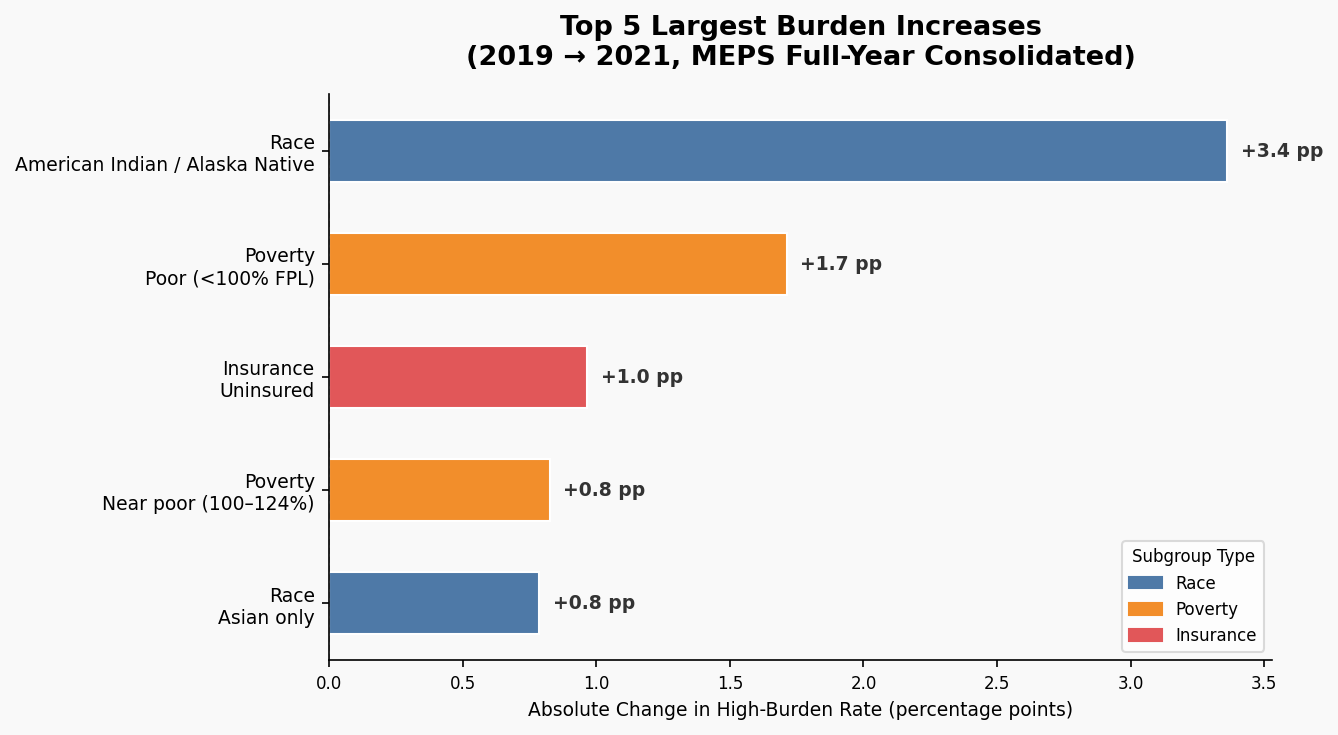}
\caption{Top five largest survey-weighted increases in high healthcare financial burden from 2019 to 2021. Changes are reported in percentage points.}
\label{fig:top5_burden_increases}
\end{figure}

Figure~\ref{fig:top5_burden_increases} presents the top five increases in high financial burden from 2019 to 2021. The biggest increases were seen among American Indian / Alaska Native persons, persons living in poverty, the uninsured, the near-poor, and Asian persons. These changes are reported in percentage points to emphasize absolute changes in burden rates, rather than relative percentage changes, which can be misleading when baseline rates are small.

Overall, the subgroup results indicate that financial vulnerability remained concentrated among already vulnerable populations in the post-pandemic period. This analysis does not show causal effects of the pandemic; however, this pattern suggests that disparities in healthcare financial burden that existed prior to the pandemic persisted and, for some groups, worsened between 2019 and 2021.

\subsection{Logistic Regression Findings}

Logistic regression was used to estimate adjusted associations between individual characteristics and high healthcare financial burden. Poverty category emerged as the strongest predictor. Relative to the high-income reference group, individuals below the federal poverty level had substantially higher odds of high financial burden, followed by near-poor, low-income, and middle-income individuals.

\begin{figure}[H]
\centering
\includegraphics[width=\textwidth]{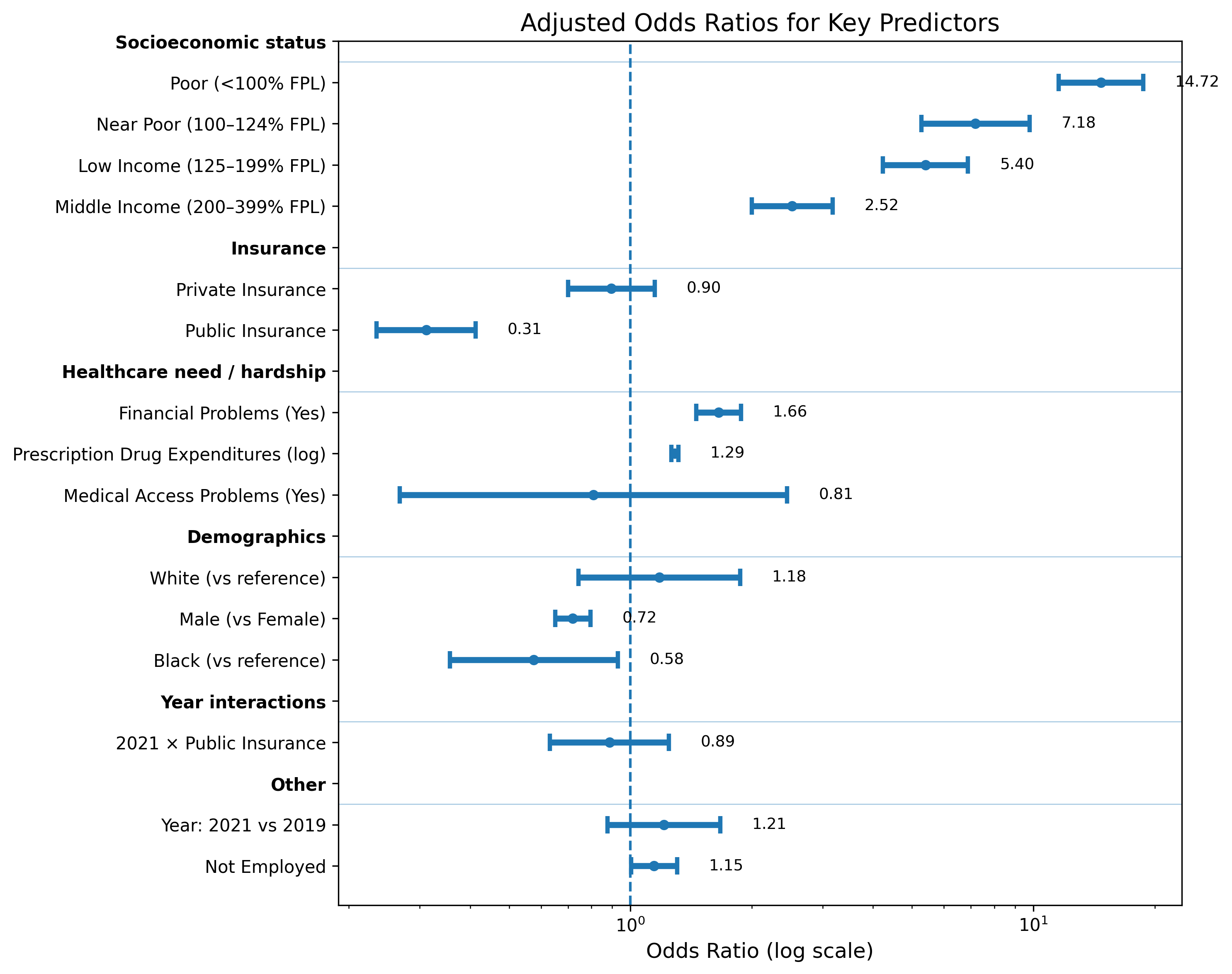}
\caption{Adjusted odds ratios (ORs) and 95\% confidence intervals from the multivariable logistic regression model. Only selected predictors with the largest effect sizes are displayed for clarity. Predictors are grouped by thematic category. The dashed vertical line indicates an odds ratio of 1.}
\label{fig:logistic_forest}
\end{figure}

As shown in Figure~\ref{fig:logistic_forest}, individuals below the federal poverty level had an adjusted odds ratio of 14.72 (95\% CI: 11.56--18.75). The corresponding odds ratios were 7.18 (95\% CI: 5.27--9.79) for near-poor individuals, 5.40 (95\% CI: 4.23--6.88) for low-income individuals, and 2.52 (95\% CI: 2.00--3.17) for middle-income individuals.

Prescription drug expenditures were positively associated with high financial burden (OR: 1.29, 95\% CI: 1.26--1.31, per one-unit increase in the log-transformed expenditure variable). Reported financial problems were also associated with higher odds of high burden (OR: 1.66, 95\% CI: 1.46--1.88). In contrast, the confidence intervals for medical access problems and inability to pay bills included an odds ratio of 1 after adjustment.

Public insurance was associated with lower odds of high financial burden relative to the reference insurance category (OR: 0.31, 95\% CI: 0.23--0.41), whereas the confidence interval for private insurance included an odds ratio of 1. These estimates should be interpreted cautiously because insurance status is also related to income, employment, health status, and healthcare utilization.

The year-by-poverty and year-by-insurance interaction estimates did not provide strong evidence that these adjusted associations differed between 2019 and 2021.

\subsection{Model Comparison}

Model comparison suggested that logistic regression, random forest, and gradient boosting all captured meaningful predictive information regarding high healthcare financial burden. However, the models had different evaluation metric trade-offs.

Gradient boosting produced the best ROC-AUC and accuracy, but its recall was low at the default classification threshold. This reflects the class imbalance in the outcome: because high financial burden is rare, the model maximizes overall accuracy by defaulting to the majority class, consequently missing most of the true high-burden cases. Logistic regression had the highest recall – it found a bigger share of high-burden individuals – but also had lower precision. The random forest model achieved the best F1 score across all the models tested, yielding a better trade-off between recall and precision.

\begin{table}[H]
\centering
\caption{Performance comparison of predictive models for predicting high healthcare financial burden.}
\label{tab:model_family_comparison}
\begin{tabular}{lccccc}
\toprule
Model & ROC-AUC & Accuracy & Recall & Precision & F1 \\
\midrule
Gradient Boosting & 0.857 & 0.968 & 0.060 & 0.541 & 0.108 \\
Random Forest & 0.840 & 0.933 & 0.413 & 0.215 & 0.283 \\
Logistic Regression & 0.838 & 0.792 & 0.743 & 0.107 & 0.186 \\
\bottomrule
\end{tabular}
\end{table}

Permutation feature importance further corroborated the importance of socioeconomic and healthcare-need variables. Across all predictive models, poverty category was the most important predictor, followed by prescription drug spending, insurance status, and age. This pattern aligns with the subgroup analysis and logistic regression results, indicating that financial vulnerability was most strongly influenced by income, insurance status, and healthcare need.

\begin{table}[H]
\centering
\caption{Permutation feature importance across predictive models. Values represent the mean decrease in ROC-AUC when each feature is permuted.}
\label{tab:permutation_importance}
\small
\begin{tabular}{lccc}
\toprule
\textbf{Feature} & \textbf{Logistic Regression} & \textbf{Random Forest} & \textbf{Gradient Boosting} \\
\midrule
POVCAT & 0.1568 & 0.1138 & 0.1161 \\
Prescription drug spending & 0.0217 & 0.0781 & 0.0966 \\
INSURC & 0.0252 & 0.0308 & 0.0308 \\
Age & 0.0844 & 0.0307 & 0.0288 \\
RACEV1X & 0.0070 & 0.0066 & 0.0035 \\
SEX & 0.0076 & 0.0064 & 0.0033 \\
\bottomrule
\end{tabular}
\end{table}

\subsection{Temporal Generalization}

To assess whether predictive relationships remained stable over time, we conducted temporal generalization experiments using the 2019 and 2021 MEPS datasets. Two within-period experiments established baseline performance by training and testing models on the same survey year, while a cross-period experiment evaluated whether a model trained on pre-pandemic data generalized to post-pandemic observations.

Table~\ref{tab:temporal_generalization} summarizes the results. The within-period experiments achieved nearly identical predictive performance, with ROC-AUC values of 0.863 for the 2019 dataset and 0.862 for the 2021 dataset. When the model trained on the 2019 dataset was evaluated on the 2021 dataset, the ROC-AUC decreased slightly to 0.846, representing a reduction of approximately 0.017.

Overall, the relatively small decline in predictive performance suggests that the relationships learned from the pre-pandemic data remained largely informative in the post-pandemic setting. This finding indicates that the principal predictors of healthcare financial vulnerability, including poverty category, insurance status, prescription drug spending, and age, retained substantial predictive value despite the disruptions associated with the COVID-19 pandemic.

\begin{table}[H]
\centering
\caption{Temporal generalization performance across pre- and post-pandemic experiments.}
\label{tab:temporal_generalization}
\small
\begin{tabular}{lccccc}
\toprule
\textbf{Experiment} & \textbf{ROC-AUC} & \textbf{Accuracy} & \textbf{Recall} & \textbf{Precision} & \textbf{F1} \\
\midrule
2019 $\rightarrow$ 2019 & 0.863 & 0.926 & 0.513 & 0.210 & 0.297 \\
2021 $\rightarrow$ 2021 & 0.862 & 0.932 & 0.443 & 0.222 & 0.296 \\
2019 $\rightarrow$ 2021 & 0.846 & 0.931 & 0.436 & 0.215 & 0.288 \\
\bottomrule
\end{tabular}
\end{table}

\section{Discussion}

We examined healthcare-related financial vulnerability before and after the onset of the COVID-19 pandemic using nationally representative MEPS data from 2019 and 2021. Across survey-weighted subgroup analyses, interpretable logistic regression, and machine learning models, the results consistently indicated that healthcare financial burden remained concentrated among socioeconomically disadvantaged populations, individuals with limited insurance protection, and those with greater healthcare-related needs.

Subgroup analyses showed increases in high healthcare financial burden between 2019 and 2021 among several already vulnerable populations, including individuals living in poverty, the uninsured, and selected racial and ethnic groups. Although the repeated cross-sectional design precludes individual-level inference or causal interpretation, the findings suggest that pre-existing disparities persisted into the post-pandemic period. These results are consistent with previous evidence that healthcare affordability is disproportionately affected by socioeconomic disadvantage and limited access to financial resources.

The logistic regression analysis reinforced the importance of socioeconomic status, with poverty category emerging as the strongest predictor of healthcare financial burden. Prescription drug expenditures and reported financial problems were also associated with increased odds of high burden, while associations for several other hardship indicators were less precise after adjustment. These findings highlight the combined importance of financial resources, insurance coverage, and healthcare need, while also showing that not every hardship-related measure contributed independently in the multivariable model.

The comparison of logistic regression, random forest, and gradient boosting demonstrated that all three approaches captured meaningful predictive information, although with different trade-offs between discrimination and interpretability. Gradient boosting achieved the highest overall discrimination, logistic regression provided the greatest interpretability and the highest recall at the evaluated classification threshold, while random forest exhibited the most balanced performance across recall, precision, and F1 score. These findings emphasize that model selection should be guided by the intended application rather than relying solely on summary performance measures such as ROC-AUC or accuracy.

A notable contribution of this study is the evaluation of temporal generalization across pre- and post-pandemic periods. Models trained on pre-pandemic data retained most of their predictive performance when evaluated on post-pandemic observations, with only a modest reduction in ROC-AUC. This suggests that the principal predictors of healthcare financial vulnerability remained relatively stable despite the healthcare and economic disruptions associated with COVID-19. At the same time, the observed decline in predictive performance highlights the importance of temporal validation before applying predictive models in changing healthcare environments.

Beyond identifying populations at elevated financial risk, this work demonstrates the value of combining interpretable statistical modeling with predictive machine learning to support population health surveillance and healthcare policy research. Such approaches may assist researchers and public health organizations in identifying vulnerable population groups, monitoring disparities over time, and informing evidence-based strategies aimed at improving healthcare affordability. Although this study did not directly evaluate digital access or telehealth utilization, these factors may represent important contextual influences on healthcare affordability and warrant further investigation.

Finally, the findings should be interpreted as descriptive and predictive rather than causal. This study does not estimate the causal effect of the COVID-19 pandemic on healthcare financial burden but instead evaluates how vulnerability patterns and predictive relationships differed between 2019 and 2021. Future work may extend this framework by incorporating additional MEPS survey years, geographic information, causal inference methods, measures of digital healthcare access, and external validation using other nationally representative datasets.

\section{Limitations}

This study has several limitations that should be considered when interpreting the findings. First, the analysis is observational and does not estimate the causal effect of the COVID-19 pandemic on healthcare financial burden. The comparison between 2019 and 2021 shows differences in population-level patterns before and after the first pandemic period, but these differences should not be interpreted as direct causal effects of the pandemic.

Second, the study follows a repeated cross-sectional design. The MEPS samples for 2019 and 2021 are different respondents, not the same people followed over time. Year-to-year differences therefore reflect differences in population-level estimates and predictive relationships, not changes in financial burden at the individual level.

Third, the analysis is constrained by the variables available in the MEPS Full-Year Consolidated files and the features available in the modeling dataset. Important contextual factors, such as availability of resources at the ZIP code level, local access to healthcare, access to broadband, availability of digital devices, and barriers to telehealth were not directly incorporated. These factors may be particularly relevant to understanding healthcare access and financial vulnerability following the pandemic.

Fourth, high financial burden was defined as out-of-pocket healthcare expenditures greater than 10\% of family income. While this expenditure-to-income threshold has been widely used in prior healthcare affordability research \cite{baird2016financial,getachew2023threshold}, different thresholds or definitions could yield different estimates of vulnerability. In addition, because the outcome is based on observed healthcare expenditures, individuals who delayed or completely avoided seeking healthcare due to financial barriers may have low or no out-of-pocket expenditures and therefore would not be classified as experiencing high financial burden under this definition. Consequently, the present analysis may underestimate financial vulnerability among populations with substantial unmet healthcare needs.

Fifth, the outcomes of predictive modeling are sensitive to the choice of features, preprocessing decisions, model specifications, and classification cut-offs. Accuracy alone can be misleading because high financial burden is relatively rare. Therefore, we report several evaluation metrics, namely ROC-AUC, recall, precision, and F1 score.

Finally, survey weights were applied for descriptive subgroup estimates to support national representativeness, while logistic regression and machine learning analyses focused on adjusted associations and predictive performance within the analytical datasets. Consequently, nationally representative interpretation is limited to the weighted descriptive results. Future work could investigate complex survey regression methods or additional validation strategies that more directly incorporate the MEPS survey design.

\section{Conclusion}

This study provides a population-level analysis of healthcare financial vulnerability in the United States using nationally representative MEPS data from 2019 and 2021. We investigated the determinants of high healthcare financial burden and the temporal stability of these relationships through survey-weighted subgroup analyses, interpretable logistic regression, machine learning models, and temporal generalization experiments.

The findings indicate that healthcare financial vulnerability remains strongly associated with socioeconomic disadvantage, insurance status, and healthcare-related need. Poverty category consistently emerged as the strongest predictor across both statistical and machine learning models, while prescription drug expenditures, insurance coverage, and age also contributed substantially to risk stratification. Importantly, the temporal generalization experiments showed only modest reductions in predictive performance when models trained on pre-pandemic data were applied to post-pandemic observations. This suggests that the principal factors associated with healthcare financial vulnerability remained relatively stable despite the healthcare and economic disruptions associated with COVID-19.

Beyond identifying vulnerable populations, this work demonstrates the value of combining interpretable statistical modeling with predictive machine learning to support population health surveillance and healthcare policy research. Such approaches may help researchers and public health organizations identify population groups at elevated risk of financial hardship, evaluate disparities across demographic and socioeconomic groups, and monitor changes in healthcare affordability over time. From a policy perspective, these findings may help policymakers, public health agencies, and healthcare systems prioritize financial assistance programs, insurance outreach initiatives, prescription assistance services, and other interventions for populations most vulnerable to healthcare-related financial hardship. The observed temporal stability also suggests that similar predictive approaches may assist in identifying financially vulnerable populations during future large-scale public health or economic disruptions, although prospective validation would be required before operational implementation.

Future work may extend this framework by incorporating additional years of MEPS data, geographic information, causal inference methods, and external validation using other nationally representative datasets. These extensions could improve our understanding of healthcare financial vulnerability and support the development of evidence-based strategies aimed at reducing financial barriers to healthcare access.

\section*{Data and Code Availability}

The Medical Expenditure Panel Survey (MEPS) data used in this study are publicly available from the Agency for Healthcare Research and Quality (AHRQ).

The code, analysis scripts, and materials necessary to reproduce the results presented in this paper are available at:

\url{https://github.com/AlexeyKresin/MEPS-Financial-Vulnerability}

\clearpage
\section*{Acknowledgments}

The authors are grateful to AI CoLab: MedStar--Georgetown Collaborative Center for Artificial Intelligence in Healthcare Research and Education for providing the research environment, mentorship, and collaborative framework that supported this work.

The authors also thank the project coordinators and administrative team, including Charlene J. Lipkins, Maryam Solimany, Omar M. Aljawfi, and Sara L. Stienecker, for their coordination, support, and contributions to the organization of the project.

\end{document}